\documentclass[runningheads]{llncs}

\usepackage[T1]{fontenc}
\usepackage{amsmath}
\usepackage[colorlinks=true]{hyperref} 
\hypersetup{
    citecolor=green
}
\usepackage{graphicx}
\usepackage{booktabs}

\usepackage{amsmath}
\usepackage{amssymb}
\begin{document}
\title{SentiMM: A Multimodal Multi-Agent Framework for Sentiment Analysis in Social Media}
\titlerunning{SentiMM}
\author{Xilai Xu\inst{1} \and
Zilin Zhao\inst{2}\and
Chengye Song\inst{3}\and
Zining Wang\inst{3}\and
Jinhe Qiang\inst{2}\and
Jiongrui Yan\inst{3}\and
Yuhuai Lin\inst{1}\thanks{Corresponding Author}
}
\authorrunning{X.Xu et al.}
\institute{College of Information and Electrical Engineering, China Agricultural University \\ \and
College of Software, Jilin University \\ \and
College of Communication Engineering, Jilin University}
%
\maketitle              
\begin{abstract}
With the increasing prevalence of multimodal content on social media, sentiment analysis faces significant challenges in effectively processing heterogeneous data and recognizing multi-label emotions. Existing methods often lack effective cross-modal fusion and external knowledge integration. We propose \textit{SentiMM}, a novel multi-agent framework designed to systematically address these challenges. \textit{SentiMM} processes text and visual inputs through specialized agents, fuses multimodal features, enriches context via knowledge retrieval, and aggregates results for final sentiment classification. We also introduce \textit{SentiMMD}, a large-scale multimodal dataset with seven fine-grained sentiment categories. Extensive experiments demonstrate that SentiMM achieves superior performance compared to state-of-the-art baselines, validating the effectiveness of our structured approach.


\end{abstract}
\section{Introduction}
Sentiment analysis is a fundamental research task in pattern recognition, aiming to automatically identify and extract subjective emotional information from text, images, or other modalities. It encompasses multiple dimensions such as opinion expression, emotion categories, and attitude tendencies. This technology has become indispensable in numerous real-world applications\cite{sa_survey}.

Over the past decade, sentiment analysis methodologies have evolved dramatically from traditional rule-based systems to data-driven machine learning approaches. The advent of Transformer-based models further propelled this progress. These deep models excel at automatically extracting complex semantic and syntactic features, achieving performance far beyond traditional methods and driving sentiment analysis to new heights.

However, with the rapid proliferation of multimedia content on social media platforms, emotional expression increasingly exhibits multi-modal and multi-label characteristics. User-generated posts often combine text, images, and audio, and a single data instance may convey multiple simultaneous emotions. The heterogeneity of modalities, emotional interference, ambiguity, and subtle contextual cues pose significant challenges to model performance and interpretability. Consequently, traditional single-modal or single-label emotion classification methods struggle to meet the demands of complex, real-world sentiment analysis \cite{multilabel,multimodal1,multimodal2}.

The emergence of Multi-modal Large Language Models (MLLMs), offers promising new directions for multimodal sentiment analysis. These models possess powerful visual perception and understanding capabilities, enabling them to process and reason over heterogeneous data. Moreover, their strong zero-shot and few-shot learning abilities allow effective sentiment classification with minimal task-specific training via prompting. Nonetheless, while MLLMs perform well on standard binary or ternary sentiment tasks, their effectiveness often diminishes on more complex scenarios requiring multi-step reasoning, structured affective inference, or integration of external knowledge.

To overcome these limitations, multi-agent systems have been proposed to handle the complexity of fine-grained sentiment classification by distributing subtasks among specialized agents. Such systems leverage complementary model strengths to boost accuracy and interpretability \cite{multiagent1}. However, challenges remain in scaling and coordination annotation costs \cite{multiagent2}.

Despite these advances, there remains an urgent need for large-scale, high-quality datasets that capture the multi-modal and multi-label nature of emotional expression on social media. Existing datasets predominantly focus on single-modal text or images \cite{xed,mms,emotic,heco}, which do not fully reflect the rich multimodal information typical of social media posts. Other works rely on facial expression analysis \cite{faba,cfaps}, but many images without visible characters also convey emotional content that should be considered.

To address these challenges, we propose \textit{SentiMM}, a modular five-stage multi-agent framework that decomposes the sentiment analysis process into distinct, collaborative components. These stages include: (1) detailed text analysis, (2) visual content analysis, (3) multimodal fusion and refinement, (4) knowledge-based context enrichment through retrieval, and (5) final sentiment classification aggregation. We design and integrate these stages into a coherent pipeline that improves both performance and interpretability.

In addition, we construct \textit{SentiMMD}, a large-scale multimodal dataset collected from social media, annotated with seven fine-grained sentiment categories. This dataset captures the complexity and diversity of emotional expression across modalities, providing a robust benchmark for multimodal sentiment analysis.

The main contributions of this study can be summarized as follows:

\begin{itemize}
    \item We design a modular five-stage multi-agent framework that combines RAG techniques to enhance multimodal sentiment analysis performance and interpretability.
    \item We construct \textit{SentiMMD}, a large-scale, carefully annotated multi-modal and multi-label sentiment dataset from social media, covering seven sentiment categories with balanced distribution.
    \item We demonstrate through extensive experiments that \textit{SentiMM} significantly outperforms state-of-the-art multimodal models on \textit{SentiMMD}, validating the effectiveness of our approach.
\end{itemize}

\begin{figure}[!t]
    \centering
    \includegraphics[width=1\linewidth]{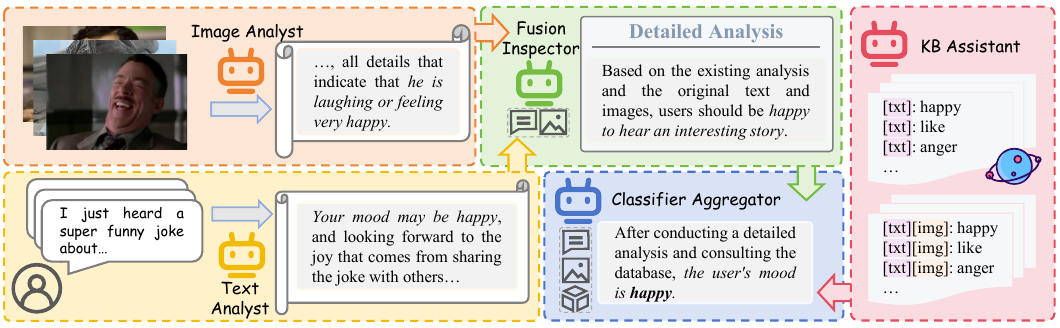}
    \caption{The overall framework of the proposed method, illustrating the main components and the data flow between each stage.}
    \label{fig:enter-label}
\end{figure}

\section{Related Work}
\subsection{Sentiment Analysis}
Sentiment analysis has traditionally focused on textual data, evolving from lexicon-based methods to sophisticated machine learning and deep learning models. Early lexicon-based approaches, such as those leveraging SentiWordNet \cite{Sentiwordnet}, assigned polarity scores to words or phrases but struggled with context and nuance. Subsequently, supervised machine learning classifiers like Support Vector Machines (SVM) and Naive Bayes \cite{svm,bayes} improved performance by learning from annotated corpora, yet they heavily depended on manual feature engineering and were limited in capturing complex linguistic phenomena.

The advent of deep learning, particularly Transformer architectures, revolutionized sentiment analysis by enabling automatic extraction of semantic and syntactic features \cite{tf4sa}. However, these models predominantly process text and often treat sentiment classification as a single-label task, which limits their applicability in scenarios involving multimodal inputs and multi-label emotions commonly found in social media.

\subsection{Multi-modal Large Language Models and Prompting Techniques}
Multi-modal large language models (MLLMs) have demonstrated strong generalization ability in sentiment classification tasks. By leveraging large-scale pre-training with multi-modal data, these models can perform zero-shot or few-shot learning through carefully designed prompts, thereby reducing the need for specific task fine-tuning.

However, MLLMs face challenges in tasks requiring integration of external knowledge. To mitigate these issues, Retrieval-Augmented Generation (RAG) frameworks combine generative models with external knowledge retrieval, enabling models to access up-to-date and domain-specific information during inference \cite{rag}. This approach reduces hallucination and improves factual accuracy, which is critical for knowledge-intensive tasks. Recent studies have applied RAG to domain-specific sentiment analysis, for example, achieving significant results in financial sentiment classification \cite{rag4sa}. However, their application in sentiment analysis, particularly multi-label emotion recognition, remains limited.

\subsection{Multi-Agent Systems in Sentiment Analysis}
Multi-agent systems decompose complex tasks into subtasks handled by specialized agents, facilitating collaboration and improving overall system performance. Wang et al. \cite{CoSA} developed Co-SA, a multi-agent system for multimodal sentiment analysis that optimizes coordination among modality-specific agents, effectively capturing cross-modal emotional cues.
Swathi et al. explored a hybrid approach that combines deep learning with multi-criteria decision-making to improve the interpretability and performance of recommendation systems \cite{hybriddeeplearning}. 

While prior works have explored multimodal sentiment analysis and multi-agent systems separately, our work uniquely integrates Retrieval-Augmented Generation within a modular multi-agent framework tailored for complex multimodal, multi-label sentiment classification on social media data.

\section{The SentiMM Framework}

\subsection{Framework Overview}

The SentiMM framework is designed based on the principles of modularity, collaboration, and hierarchical processing. It consists of five main stages, where each stage is responsible for a specific aspect of multimodal sentiment analysis. The input multimodal data is processed sequentially through these stages, with the output of each stage serving as the input for the next.

Mathematically, let the input multimodal data be represented as \(\mathbf{D} = \{T, I, V\}\), where \(T\) represents text data, \(I\) represents image data, and \(V\) represents video data. The output of the SentiMM framework is the final sentiment classification result \(C\), which belongs to one of the seven predefined categories:

\begin{equation}
\mathcal{C} = \{\text{Like}, \text{Happiness}, \text{Anger}, \text{Disgust}, \text{Fear}, \text{Sadness}, \text{Surprise}\}.
\end{equation}

The overall process of the SentiMM framework can be described as a series of operations \(\mathcal{O}_1, \mathcal{O}_2, \mathcal{O}_3, \mathcal{O}_4, \mathcal{O}_5\) corresponding to the five stages:

\begin{equation}
C = \mathcal{O}_5\big(\mathcal{O}_4\big(\mathcal{O}_3\big(\mathcal{O}_2\big(\mathcal{O}_1(\mathbf{D})\big)\big)\big)\big).
\end{equation}

In the following sections, we elaborate on each stage of the SentiMM framework in detail, using both textual descriptions and mathematical formulations to illustrate the underlying algorithms and processes.

\begin{figure}[!t]
    \centering
    \includegraphics[width=1\linewidth]{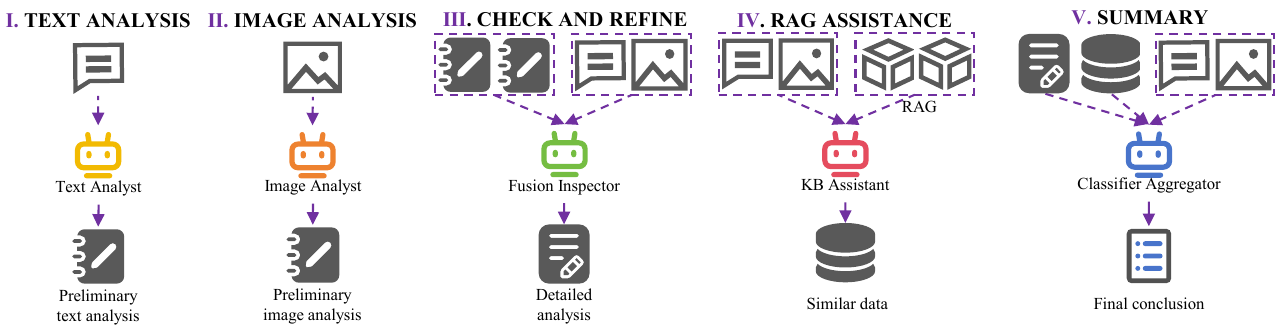}
    \caption{The five-stage pipeline of the SentiMM framework.}
    \label{fig:label2}
\end{figure}

\textit{SentiMM} comprehensively simulates the entire multimodal sentiment analysis process through a structured five-stage pipeline, as illustrated in Figure~\ref{fig:label2}. Each stage processes diverse inputs—including predicted numerical sentiment scores, detailed textual analyses, and multimodal data—and produces comprehensive outputs that serve as inputs for subsequent stages, enabling a thorough and nuanced understanding of sentiment across modalities.

\subsection{Step 1: Text Analyst - Text Analysis}

The initial stage of the \textit{SentiMM} framework is the \textit{Text Analyst} agent, which leverages a large language model (LLM) to perform an in-depth analysis of the textual component of the multimodal input. Textual data in social media and other user-generated content often encapsulates complex semantic structures, emotional nuances, and implicit sentiment cues conveyed through idiomatic expressions, sarcasm, or contextual subtleties. The Text Analyst’s objective is to extract these rich emotional signals and represent them both quantitatively and qualitatively.

Formally, the input to this stage is the raw text data \(T\). Prior to analysis, the text undergoes a rigorous preprocessing pipeline, including tokenization, stop-word removal, and lemmatization, yielding a cleaned and normalized text representation:

\begin{equation}
    T_{\text{pre}} = \text{Preprocess}(T).
\end{equation}

The Text Analyst then segments \(T_{\text{pre}}\) into \(n\) meaningful units (e.g., sentences or clauses), denoted as \(\{s_i\}_{i=1}^n\), and applies the LLM-based model \(M_{\text{text}}\) to each segment to extract sentiment scores and semantic features:

\begin{equation}
    (s_{\text{score}}(s_i), f_{\text{text}}(s_i)) = M_{\text{text}}(s_i),
\end{equation}

where \(s_{\text{score}}(s_i) \in [-1,1]\) quantifies the sentiment polarity (from extremely negative to extremely positive), and \(f_{\text{text}}(s_i) \in \mathbb{R}^d\) is a high-dimensional semantic feature vector capturing contextual and emotional nuances.

To aggregate segment-level sentiment into an overall textual sentiment score \(S_{\text{text}}\), a weighted sum is computed:

\begin{equation}
    S_{\text{text}} = \sum_{i=1}^n w_i s_{\text{score}}(s_i), \quad \text{with} \quad \sum_{i=1}^n w_i = 1,
\end{equation}

where weights \(w_i\) reflect the relative importance of each segment, determined by factors such as semantic relevance or segment length.

The output of the Text Analyst is thus a composite object comprising the overall sentiment score, the set of segment-level sentiment scores, and the semantic feature representations:

\begin{equation}
    O_{\text{text}} = \left\{ S_{\text{text}}, \{s_{\text{score}}(s_i)\}_{i=1}^n, \{f_{\text{text}}(s_i)\}_{i=1}^n \right\}.
\end{equation}

This rich output not only provides a numerical sentiment estimate but also detailed textual insights that will be crucial for multimodal fusion.

\subsection{Step 2: Image Analyst - Image Analysis}

The second stage, the \textit{Image Analyst}, is responsible for analyzing the visual modality, which may consist of static images or dynamic video sequences. Visual content conveys sentiment through facial expressions, gestures, color schemes, and scene context, all of which require sophisticated interpretation.

The input to this stage is either an image \(I\) or a video \(V\).

For video inputs, the video is first decomposed into a sequence of \(m\) frames:

\begin{equation}
    I_f = \{I_{f_k}\}_{k=1}^m = \text{Decompose}(V).
\end{equation}

For each image frame \(I_j\) (either a single image or a video frame), the Image Analyst extracts visual features using a convolutional neural network model \(M_{\text{image}}\):

\begin{equation}
    F_{\text{vis}}(I_j) = M_{\text{image}}(I_j) \in \mathbb{R}^p,
\end{equation}

where \(p\) is the dimensionality of the visual feature space.

These features are then input into a sentiment prediction sub-model \(M_{\text{senti-image}}\), which integrates the CNN features with semantic understanding from the LLM to produce a sentiment score and a detailed visual sentiment report:

\begin{equation}
    (S_{\text{image}}(I_j), R_{\text{image}}(I_j)) = M_{\text{senti-image}}(F_{\text{vis}}(I_j)),
\end{equation}

where \(S_{\text{image}}(I_j) \in [-1,1]\) is the sentiment score for image \(I_j\), and \(R_{\text{image}}(I_j)\) is a textual or structured report describing detected emotional cues.

For video inputs, the overall video sentiment score \(S_{\text{video}}\) and aggregated report \(R_{\text{video}}\) are computed by weighted aggregation over frames:

\begin{equation}
    S_{\text{video}} = \sum_{k=1}^m w_k S_{\text{image}}(I_{f_k}), \quad R_{\text{video}} = \text{AggregateReports}(\{R_{\text{image}}(I_{f_k})\}_{k=1}^m),
\end{equation}

with weights \(w_k\) reflecting frame importance based on saliency or temporal relevance.

The output of the Image Analyst is thus:

\begin{equation}
    O_{\text{image}} = 
    \begin{cases}
        \{S_{\text{image}}(I), R_{\text{image}}(I)\}, & \text{if input is image } I, \\
        \{S_{\text{video}}, R_{\text{video}}\}, & \text{if input is video } V.
    \end{cases}
\end{equation}

This output encapsulates both quantitative sentiment scores and qualitative visual sentiment descriptions, providing a comprehensive analysis of the visual modality.

\subsection{Step 3: Fusion Inspector — Check and Refine}

The third stage, the \textit{Fusion Inspector}, is tasked with integrating the textual and visual sentiment analyses to produce a unified multimodal sentiment representation. Beyond simple fusion, this stage emphasizes checking for inconsistencies or conflicts between modalities and refining the sentiment estimation by addressing such discrepancies.

The inputs to this stage are the original text \(T\), the image or video data \(I\) or \(V\), and the outputs from the previous two stages:

\begin{equation}
    \left( T, I \text{ or } V, O_{\text{text}}, O_{\text{image}} \right).
\end{equation}

From \(O_{\text{text}}\) and \(O_{\text{image}}\), we extract the semantic feature sets:

\begin{equation}
    F_{\text{text}} = \{f_{\text{text}}(s_i)\}_{i=1}^n, \quad F_{\text{vis}} = 
    \begin{cases}
        F_{\text{vis}}(I), & \text{if image}, \\
        \{F_{\text{vis}}(I_{f_k})\}_{k=1}^m, & \text{if video}.
    \end{cases}
\end{equation}

These features are combined into a joint multimodal feature vector \(F_{\text{combined}}\) via concatenation or a learned fusion function \(\phi\):

\begin{equation}
    F_{\text{combined}} = \phi(F_{\text{text}}, F_{\text{vis}}).
\end{equation}

The fusion model \(M_{\text{fusion}}\) processes \(F_{\text{combined}}\) to produce a multimodal sentiment score and a detailed fusion report:

\begin{equation}
    (S_{\text{multimodal}}, R_{\text{fusion}}) = M_{\text{fusion}}(F_{\text{combined}}),
\end{equation}

where \(S_{\text{multimodal}} \in [-1,1]\) and \(R_{\text{fusion}}\) contains interpretive insights on cross-modal sentiment alignment or conflict.

To check for inconsistencies or missing information, difference metrics \(\Delta S_i\) are computed between the multimodal sentiment and each unimodal sentiment score:

\begin{equation}
    \Delta S_{\text{text}} = |S_{\text{multimodal}} - S_{\text{text}}|, \quad
    \Delta S_{\text{image}} = 
    \begin{cases}
        |S_{\text{multimodal}} - S_{\text{image}}(I)|, & \text{if image}, \\
        |S_{\text{multimodal}} - S_{\text{video}}|, & \text{if video}.
    \end{cases}
\end{equation}

If any \(\Delta S_i\) exceeds a predefined threshold \(\theta\), the Fusion Inspector \textbf{refines} the analysis by invoking an auxiliary analysis module \(M_{\text{aux}}\) to hypothesize missing or conflicting sentiment cues:

\begin{equation}
    H = M_{\text{aux}}(T, I \text{ or } V, O_{\text{text}}, O_{\text{image}}, S_{\text{multimodal}}),
\end{equation}

where \(H\) is a set of textual hypotheses or flagged inconsistencies that guide further interpretation.

The output of this stage is a comprehensive package:

\begin{equation}
    O_{\text{fusion}} = \{ S_{\text{multimodal}}, R_{\text{fusion}}, \Delta S_{\text{text}}, \Delta S_{\text{image}}, H \}.
\end{equation}

This output provides a robust foundation for subsequent context enrichment and final decision-making.

\subsection{Step 4: KB Assistant — Enrich Context}

The fourth stage, the \textit{KB Assistant}, employs a Retrieval-Augmented Generation (RAG) system to enrich the sentiment analysis with external knowledge. By retrieving semantically similar historical multimodal data and their sentiment annotations, this stage provides additional context that enhances robustness and interpretability.

Inputs to this stage include the original multimodal data and the fusion output:

\begin{equation}
    \left( T, I \text{ or } V, O_{\text{fusion}}, D_{\text{rag}} \right),
\end{equation}

where \(D_{\text{rag}}\) is the RAG database containing indexed multimodal entries with sentiment annotations.

Key features \(F_{\text{key}}\) are extracted from the fusion features using a selection function \(\psi\) guided by the LLM:

\begin{equation}
    F_{\text{key}} = \psi(F_{\text{combined}}).
\end{equation}

The RAG retrieval mechanism computes similarity scores between \(F_{\text{key}}\) and each database entry \(d_i \in D_{\text{rag}}\):

\begin{equation}
    \text{sim}(d_i, F_{\text{key}}) = \text{SimilarityMeasure}(d_i, F_{\text{key}}).
\end{equation}

The top-\(k\) most similar entries are selected:

\begin{equation}
    D_{\text{similar}} = \text{Top-}k\left(\{\text{sim}(d_i, F_{\text{key}})\}_{i=1}^{|D_{\text{rag}}|}\right).
\end{equation}

Each retrieved entry \(d_j \in D_{\text{similar}}\) includes multimodal data and associated sentiment annotations \(S(d_j)\), as well as metadata \(M(d_j)\).

The KB Assistant processes \(D_{\text{similar}}\) to generate a contextual report \(R_{\text{rag}}\) that summarizes relevant patterns, sentiment trends, and domain-specific insights:

\begin{equation}
    R_{\text{rag}} = M_{\text{rag}}(D_{\text{similar}}).
\end{equation}

The output of this stage is:

\begin{equation}
    O_{\text{rag}} = \{ D_{\text{similar}}, R_{\text{rag}} \}.
\end{equation}

This enriched context will be instrumental in refining the final sentiment classification.

\subsection{Step 5: Classifier Aggregator — Summary}

The final stage, the \textit{Sentiment Classifier Aggregator}, synthesizes all prior outputs to generate a definitive sentiment classification. It integrates the multimodal sentiment score, retrieved similar data, and the original multimodal inputs to produce a nuanced, context-aware sentiment label.

The inputs are:

\begin{equation}
    \left( O_{\text{fusion}}, O_{\text{rag}}, T, I, V \right).
\end{equation}

From the retrieved similar data \(D_{\text{similar}}\), an average sentiment score \(S_{\text{similar}}\) is computed:

\begin{equation}
    S_{\text{similar}} = \frac{1}{|D_{\text{similar}}|} \sum_{d_j \in D_{\text{similar}}} S(d_j).
\end{equation}

The aggregator combines the multimodal sentiment score \(S_{\text{multimodal}}\) and the retrieved sentiment \(S_{\text{similar}}\) via a weighted sum:

\begin{equation}
    S_{\text{combined}} = \alpha S_{\text{multimodal}} + \beta S_{\text{similar}}, \quad \text{with} \quad \alpha + \beta = 1,
\end{equation}

where \(\alpha\) and \(\beta\) are hyperparameters tuned to balance direct analysis and contextual knowledge.

The combined sentiment score \(S_{\text{combined}}\) is then input to a classification model \(M_{\text{classify}}\), which also considers the original multimodal data and auxiliary reports to assign a final sentiment label \(L\) from a predefined set \(\mathcal{L}\):

\begin{equation}
\begin{split}
    L &= M_{\text{classify}}(S_{\text{combined}}, R_{\text{fusion}}, R_{\text{rag}}, T, I, V), \\
    L &\in \mathcal{L} = \{\text{Like, Happiness, Anger, Disgust, Fear, Sadness, Surprise}\}.
\end{split}
\end{equation}

The output of this stage is the final sentiment classification along with an explanatory report \(R_{\text{final}}\) that justifies the decision based on multimodal evidence and retrieved context:

\begin{equation}
    O_{\text{final}} = \{ L, R_{\text{final}} \}.
\end{equation}

This comprehensive output concludes the \textit{SentiMM} pipeline, providing both a categorical sentiment label and a rich interpretive summary.

\section{The SentiMMD Dataset}
\subsection{Data Collection}
The SentiMMD dataset is constructed by crawling multimodal data from various social media platforms, including posts containing images with accompanying text, as well as online videos paired with their titles or descriptions. For video data, multiple keyframes are extracted to represent the visual content effectively. Each data instance thus consists of at least one textual modality and one visual modality (image or video frames).

To ensure high-quality annotations, each sample is labeled by 40 human annotators who assign one of seven sentiment labels: \textit{Like, Happiness, Anger, Disgust, Fear, Sadness, Surprise}. Multiple rounds of manual verification are conducted to guarantee consistency between human annotations and GPT-based sentiment predictions. For samples where human and GPT annotations disagree, expert reviewers perform re-annotation or exclude ambiguous samples to maintain dataset integrity.

\subsection{Data Statistics}

The dataset contains a total of 3,500 multimodal samples, evenly distributed across the seven sentiment categories, with 500 samples per label. The dataset is split into training and testing sets, where 10\% of the data (350 samples) is reserved as the test set, and the remaining 90\% (3,150 samples) is used for training.
Textual data averages 28.7 tokens per entry, with image resolution standardized to 224×224 pixels. Video data has an average duration of 12.3 seconds, yielding 3-7 keyframes per video after preprocessing.

\subsection{Knowledge Base}
The knowledge base (KB) constructed for the SentiMMD dataset plays a pivotal role in enhancing the multimodal sentiment analysis process by providing rich, domain-specific contextual information. It is designed as a comprehensive repository that stores not only the multimodal training data but also a large collection of traditional pure-text sentiment evaluation datasets collected from the web. This hybrid KB enables the SentiMM framework to leverage both multimodal and textual knowledge sources, facilitating more accurate and context-aware sentiment classification.

Specifically, the KB contains:

\begin{itemize}
    \item \textbf{Multimodal Training Data:} All training samples from the SentiMMD dataset, including text, images, videos, and their corresponding sentiment labels. This ensures that the KB is tightly coupled with the target domain and task.
    \item \textbf{Traditional Textual Sentiment Datasets:} Curated datasets from publicly available sentiment analysis benchmarks, and Twitter sentiment corpora. These datasets provide a rich source of linguistic sentiment patterns and lexical cues.

    \end{itemize}

The KB is indexed using a vector database optimized for fast nearest neighbor search, allowing the KB Assistant module in the SentiMM framework to retrieve semantically relevant examples that enrich the sentiment context of the input data. 
By integrating this knowledge base, the SentiMM framework benefits from a robust external memory that complements the large language models’ reasoning capabilities.

\section{Experiments}

\subsection{Experimental Settings}

All experiments are conducted using large multimodal models, primarily GPT-4o and Qwen2.5-VL-7B, which serve as the backbone for various stages of the SentiMM framework. The models are prompted appropriately to handle multimodal inputs and sentiment classification tasks.

Evaluation metrics include accuracy (Acc.), macro F1-score (MF), and confusion matrix analysis to assess classification performance across all seven sentiment categories.

\subsection{Baselines}

We compare the SentiMM framework against several strong baselines, including:

\begin{itemize}
    \item \textbf{GLM-4V-9B} \cite{glm2024chatglmfamilylargelanguage}: A large-scale multimodal language model consisting of 9 billion parameters, designed to handle both visual and textual inputs.
    \item \textbf{mPLUG-7B} \cite{ye2024mplugowlmodularizationempowerslarge}: A 7 billion parameter multimodal pre-trained model that integrates vision and language understanding through modular design.
    \item \textbf{Qwen2.5-VL-3B} \cite{qwen2.5vl}: A compact 3 billion parameter variant of the Qwen multimodal model, optimized for efficient vision-language tasks.
    \item \textbf{Qwen2.5-VL-7B}: The full-scale 7 billion parameter Qwen multimodal model, offering enhanced capacity for complex multimodal understanding.
    \item \textbf{GPT-4o-20250513} \cite{openai2024gpt4ocard}: A state-of-the-art multimodal large language model, supporting a wide range of vision-language tasks with high accuracy and fluency.
\end{itemize}

All baseline models are fine-tuned on the same training split of the SentiMMD dataset and evaluated on the identical test set using consistent metrics, ensuring a fair and direct comparison.

\subsection{Overall Performance}
Table~\ref{tab:overall_performance} presents the comprehensive evaluation results of the \textit{SentiMM} framework against baseline models on the SentiMMD test set. The results demonstrate a substantial performance gap between \textit{SentiMM} and all baseline models, validating the effectiveness of the proposed multimodal sentiment analysis pipeline.

As shown, the \textit{SentiMM} framework significantly outperforms all baseline models by a large margin, achieving an accuracy of 89.3\% with GPT-4o as the backbone. This demonstrates the effectiveness of the modular five-stage pipeline and the integration of the knowledge base in enhancing multimodal sentiment understanding. Notably, the macro F1-score (MF1) of 88.5\% indicates balanced performance across sentiment classes, reflecting robustness in handling class imbalance. Compared to the strongest baseline GPT-4o, \textit{SentiMM} improves accuracy by 11.5 percentage points, which is a substantial gain in the challenging multimodal sentiment analysis task.

\begin{table}[!t]
\centering
\caption{Overall performance comparison of \textit{SentiMM} and baseline models on the SentiMMD test set. Metrics reported are accuracy (\%), macro precision (MP), macro recall (MR), and macro F1-score (MF1).}
\label{tab:overall_performance}
\begin{tabular}{lcccc}
\toprule
\textbf{Model} & \textbf{Acc. (\%)} & \textbf{MP (\%)} & \textbf{MR (\%)} & \textbf{MF1 (\%)} \\
\midrule
GLM-4V-9B & 68.2 & 67.5 & 66.8 & 67.1 \\
mPLUG-7B & 65.4 & 64.7 & 63.9 & 64.3 \\
Qwen2.5-VL-3B & 59.7 & 57.1 & 58.2 & 57.6 \\
Qwen2.5-VL-7B & 75.1 & 74.5 & 73.9 & 74.2 \\
GPT-4o-20250513 & 77.8 & 77.2 & 76.7 & 77.0 \\
\midrule
\textbf{SentiMM (GPT-4o)} & \textbf{89.3} & \textbf{88.7} & \textbf{88.1} & \textbf{88.4} \\
\textbf{SentiMM (Qwen2.5-VL-7B)} & 82.1 & 81.5 & 81.0 & 81.2 \\
\bottomrule
\end{tabular}
\end{table}

\subsection{Qualitative Analysis}

To further understand the contributions of each component in the SentiMM framework, we conduct ablation studies by systematically removing or modifying key modules. Table~\ref{tab:ablation_study} summarizes the results.
\begin{table}[!t]
\centering
\caption{Ablation study results on the SentiMMD test set. Metrics reported are accuracy (\%) and macro F1-score (MF1, \%). The baseline is the full \textit{SentiMM} framework with GPT-4o.}
\label{tab:ablation_study}
\begin{tabular}{lcc}
\toprule
\textbf{Model Variant} & \textbf{Accuracy (\%)} & \textbf{MF1 (\%)} \\
\midrule
Full \textit{SentiMM} (GPT-4o) & 89.3 & 88.4 \\
\quad w/o KB Assistant (no retrieval) & 84.7 & 83.9 \\
\quad w/o Fusion Inspector (no refinement) & 85.2 & 83.7 \\
\quad w/o Image Analyst (text only) & 78.9 & 77.8 \\
\quad w/o Text Analyst (image/video only) & 74.3 & 73.5 \\
\quad w/o Classifier Aggregator (direct fusion only) & 83.5 & 81.8 \\
\bottomrule
\end{tabular}
\end{table}

Removing the Fusion Inspector module, which refines the multimodal fusion results, causes the accuracy to drop from 89.3\% to 85.2\%. This indicates that refinement is crucial for resolving conflicts and enhancing feature integration.

When the image analyst is removed, leaving only text input, performance drops sharply by 10.4 points in accuracy. This confirms that visual information is indispensable for accurate multimodal sentiment analysis.

Similarly, removing the text analyst and relying solely on image/video input results in the largest performance degradation, showing that textual cues are even more critical for sentiment understanding in this dataset.

The classifier aggregator combines multiple classifier outputs for final decision making. Without it, the accuracy drop by about 5.8 points, demonstrating the effectiveness of ensemble strategies in boosting robustness.

Overall, the ablation results validate the design choices of the \textit{SentiMM} framework. Each module contributes significantly to the final performance, and the synergy among them leads to state-of-the-art results.

\section{Conclusion}
We have presented \textit{SentiMM}, a novel multi-agent framework for multimodal, multi-label sentiment analysis on social media data. By decomposing the task into specialized stages—text analysis, image analysis, fusion inspection, knowledge enrichment, and classification aggregation—\textit{SentiMM} achieves superior performance and interpretability. Complementing this, we introduce \textit{SentiMMD}, a large-scale, carefully annotated multimodal sentiment dataset covering seven fine-grained emotions. Extensive experiments demonstrate that \textit{SentiMM} significantly outperforms state-of-the-art baselines, validating the benefits of modular design and external knowledge integration. Future work will explore extending the framework to additional modalities and real-time sentiment tracking.

\bibliographystyle{splncs04}
\bibliography{referencesflie}
\end{document}